\documentclass{article}
\usepackage{iclr2027_conference,times}
\iclrfinalcopy

\usepackage[T1]{fontenc}
\usepackage[utf8]{inputenc}
\usepackage{amsmath,amssymb}
\usepackage{xcolor}
\definecolor{teal}{HTML}{00796B}
\definecolor{violet}{HTML}{6A1B9A}
\newcommand{\reopened}[1]{\textcolor{gray}{\((-#1)\)}}
\usepackage{graphicx}
\usepackage{subcaption}
\usepackage{algorithm}
\usepackage{algpseudocode}
\algrenewcommand\algorithmicprocedure{}
\algrenewcommand\algorithmicend{\textbf{end}}
\usepackage{microtype}
\usepackage{url}
\usepackage[hidelinks]{hyperref}
\hypersetup{
  pdftitle={From Latents to Wires: Surgical Post-Editing on Large Language
  Models},
  pdfauthor={Jiankai Jin, Xiangzheng Zhang, Zhao Liu, Wenzhuo Xu,
  Dongdong Yang, Deyue Zhang, Quanchen Zou} }

\begin{document}
\title{From Latents to Wires: Surgical Post-Editing \\ on Large Language Models}
\author{%
  \begin{minipage}{0.96\textwidth}
  \centering
  \normalfont
  \begin{tabular}{@{}ccc@{}}
  \textbf{Jiankai Jin} & \textbf{Xiangzheng Zhang} & \textbf{Zhao Liu}\\
  {\small\ttfamily jinjiankai@360.cn} &
  {\small\ttfamily zhangxiangzheng@360.cn} &
  {\small\ttfamily liuzhao3@360.cn}\\[0.9em]
  \textbf{Wenzhuo Xu} & \textbf{Dongdong Yang} & \textbf{Deyue Zhang}\\
  {\small\ttfamily xuwenzhuo@360.cn} &
  {\small\ttfamily yangdongdong1@360.cn} &
  {\small\ttfamily zhangdeyue@360.cn}\\[0.9em]
  \multicolumn{3}{c}{\textbf{Quanchen Zou}}\\
  \multicolumn{3}{c}{\small\ttfamily zouquanchen@360.cn}
  \end{tabular}\\[0.8em]
  360 AI Security Lab
  \end{minipage}%
}
\maketitle
\fancyhead{}
\renewcommand{\headrulewidth}{0pt}

\begin{abstract}
Given a large language model (LLM), can whoever holds the weights name a
semantic target (e.g., the model's identity), locate the model components that
produce it, and edit them so that the target no longer appears while other
capability is preserved? We call such an edit on a trained model a post-edit. We
present L2W (latents to wires), a framework that performs surgical post-edits
for named semantic targets. For localization, L2W uses Jacobian lens (J-lens)
attribution to score components against the semantic target. For surgical
removal, because LLM mechanisms are redundant (i.e., a semantic target may have
multiple components producing it), L2W runs Counterexample-Guided Causal Cut
(CGCC) until the target no longer appears. CGCC first cumulatively closes model
components, treating each surviving expression of the target as a counterexample
that exposes the next components to close, and then reopens some of them to
preserve capability. In a controlled experiment with an implanted
behavioural watermark, L2W removes the watermark, and its localization lands on
the model region the implant changed. Across three model configurations, L2W removes
model-metadata (e.g., identity) self-claims in all nine runs, and
adult-content refusal in all three, with no held-out target residual. L2W
further composes two post-edits on a text-to-image model: one removes the refusal of requested
nudity, and a second removes the nude rendering the first exposes. The results
support post-editing as a complement to post-training: post-training installs
preferred behaviours, and post-editing removes named unwanted ones.
\end{abstract}

\section{Introduction}
\label{sec:intro}

A large language model (LLM) has typically been treated as a fixed artifact:
once trained, aligned, and evaluated, it is deployed with its weights final.
Open-weight releases change this: for
whoever holds the weights, the model is an editable program whose components
(e.g., attention heads and MLP channels) can be altered. Editing could overwrite
components with new weights, but its effect is hard to predict.
Zeroing a component's output instead, which we call closing the component, has
an exact effect: whatever the component produced is gone. The price is that such
an edit can only remove. However, when we aim to remove a semantic target (e.g., 
a static memory such as the model's identity, or a behaviour such as refusing a
sensitive request), closing is the only edit we need. The hurdle is localization:
finding a set of components whose closure stops the target from appearing.

We call such a targeted intervention a post-edit: as post-training is training
applied to a trained model, a post-edit is an edit applied to one.
The two naturally compose: a model post-trained to install
preferred behaviours can be post-edited to remove unwanted ones.

Removing a target by editing a small part of the model is no longer speculative:
\citet{arditi2024refusal} show that refusal across many chat models is mediated
by a residual direction, which a weight edit ablates; a wave of follow-up work
supports the same finding \citep{joad2026refusal,zhao2025harmfulness,siu2026repit,
chen2025safetyneurons,wu2026neurostrike,bair2026compressed}.
However, these methods share a limitation: each fixes its target first and
builds the analysis around it; none takes a target named in text as an input. The challenge,
then, is the general case: given an ordinarily trained model, can an operator
name a semantic target, find the specific model components that produce it, and
close them so that the target no longer appears while other capability is
preserved? For example, an operator may want to remove an identity claim the
model makes, or a dangerous capability it retains.

\begin{figure}[t]
  \centering
  \includegraphics[width=\linewidth]{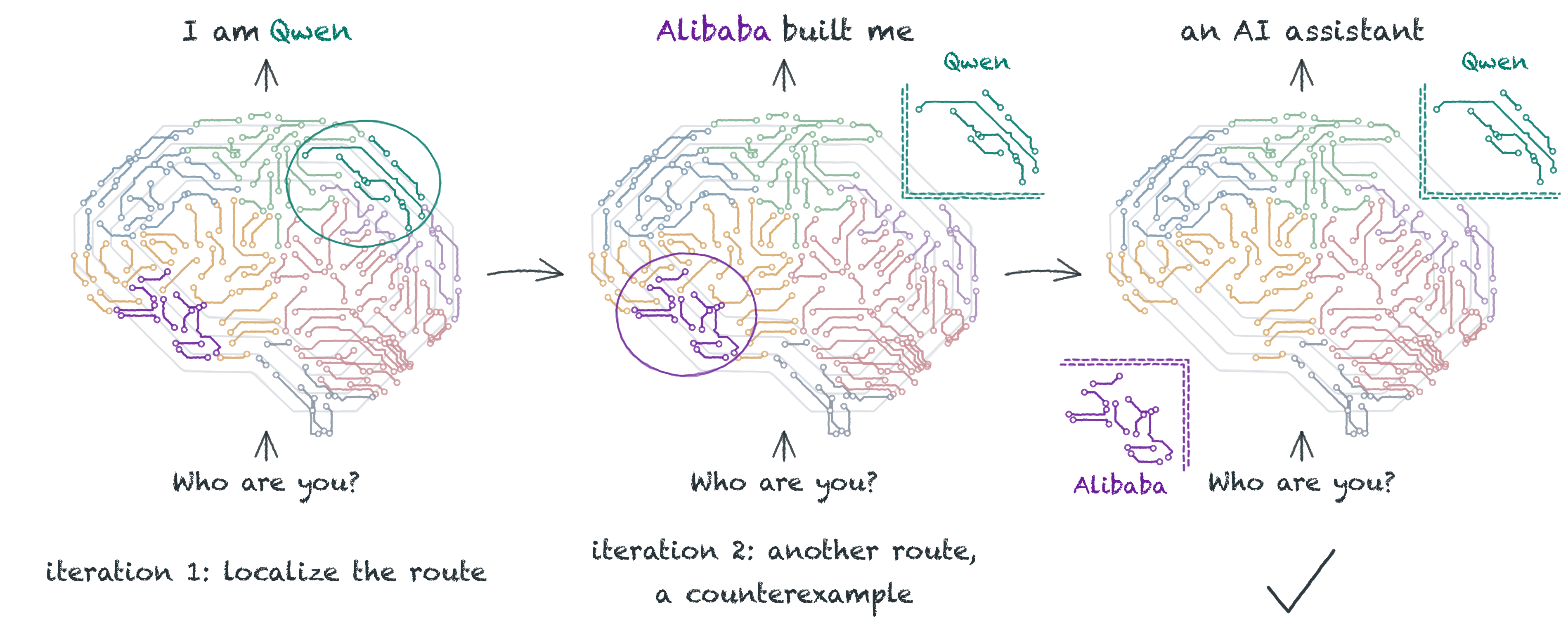}
  \caption{Surgical post-editing of an identity claim: three states of one model
  under one test probe. Left: localization finds the components producing
  \texttt{\textcolor{teal}{Qwen}}. Middle: those components are closed; the
  target survives through other components and the answer drifts to
  \texttt{\textcolor{violet}{Alibaba}}, a counterexample that drives the next
  iteration. Right: with both component groups closed, the identity claim
  disappears. }
  \label{fig:mechanism}
\end{figure}

We present L2W (latents to wires) to answer this challenge. L2W uses the
Jacobian lens (J-lens) of \citet{gurnee2026workspace} to localize the model
components producing the semantic target, and closes them as a surgical
post-edit (Figure~\ref{fig:mechanism}, data from
Section~\ref{sec:model-metadata-removal}). A single closure is rarely the end,
because LLM mechanisms are redundant: the target often re-emerges through other
components, changing how it is expressed. L2W therefore removes iteratively with
Counterexample-Guided Causal Cut (CGCC): every response that still expresses the
target is a counterexample that exposes the next components to close. Once the
target no longer appears, CGCC's restoration pass reopens the closures the
removal does not need, returning the capability they cost.

The experiments in Section~\ref{sec:experiments} support the design. In a
controlled setting where we implant a watermark and know its mechanism, the cut
lands on the chokepoint of the region the implant changed, demonstrating that
the post-edit is surgical. Across models and targets, most removals
complete with capability intact. Our contributions are:
\begin{enumerate}

\item A formulation of post-editing: removal of a named semantic target from a
  trained model. The target arrives as a natural-language description, and the
  edit closes components producing it, while other capability is preserved.

\item L2W, a model post-editing framework: J-lens localization nominates
  candidate components producing the semantic target, CGCC iterates for
  component closure until no target remains, and its restoration pass reopens
  what the removal does not need.

\item Experimental results on removal of an implanted watermark,
  removal of model-metadata self-claims and adult-content refusal across three
  model configurations, and composition of two sequential post-edits on the text encoder of a
  text-to-image model demonstrate the effectiveness of L2W and the potential of
  post-editing to complement post-training.
\end{enumerate}

\section{Related Work}
\label{sec:related}

\paragraph{Post-hoc component localization.}

The Jacobian lens of \citet{gurnee2026workspace} derives, from the model's own
weights, a direction for every vocabulary token at every layer; projecting a
layer's activations onto that direction measures the token's expression per
position. Gurnee et al.\ introduce the lens to study J-space; L2W
uses it to locate components producing a semantic target.

Very few neurons carry most of a model's alignment effect, and pruning fewer
than 0.6\% of them defeats it \citep{chen2025safetyneurons,wu2026neurostrike};
\citet{prakash2025dissecting} find the semantic re-routing our CGCC handles
dynamically: ablating one refusal feature reactivates overlapping features that
preserve the behaviour. These studies fix their target in advance and build the
analysis around it. Circuit discovery lifts that restriction, accepting any task
expressed as a metric over curated input pairs (e.g., refusal and non-refusal
inputs) \citep{wang2023ioi,goldowskydill2023path,
chan2022scrubbing,conmy2023acdc,syed2024attribution}. L2W keeps the two steps of
circuit discovery, a cheap gradient estimate that proposes components and a
causal test that confirms them, and differs at both ends. The task is a target
the operator names in text; the output is not a subgraph that explains the
behaviour but a cut that removes it. \citet{bair2026compressed} are our closest
neighbour: they localize general capabilities by knocking out random attention
heads and recovering per-head contributions as a sparse linear inverse problem.
Our method instead uses J-lens attribution to nominate candidate components for
named targets.

\paragraph{Model editing and unlearning.}

Model knowledge editing rewrites memorized facts through weight updates at the
layers causal tracing selects or through trained auxiliary editors
\citep{meng2022rome,meng2023memit,mitchell2022mend,mitchell2022serac}.
\citet{hase2023localization} report that where causal tracing places a fact
barely predicts where an edit succeeds. Our localization is from a different
perspective: the components that produce the target at inference are where an
edit is needed.

Unlearning pursues the same end state, a target that no longer surfaces while
other capability survives, but through training that updates a wide range of
parameters \citep{maini2024tofu,li2024wmdp,eldan2023harrypotter}, and the
suppressed behaviour often survives underneath
\citep{jain2024mechanistically,lee2024mechanistic,qi2025safety}; L2W provides an
explicit cut. Directional ablation reaches the same end state at inference: it
fits a refusal direction from contrasting probe sets and projects it out of the
residual stream at every layer \citep{arditi2024refusal};
Section~\ref{sec:arditi} compares it with L2W under a matched probe budget.
Watermark and backdoor removal is usually a training defence, washing the
implant out by fine-tuning and pruning \citep{liu2020wild,liu2018finepruning};
L2W instead locates the implanted region and closes it.
Structured pruning also closes components, but to shrink the model with
capability mostly preserved \citep{michel2019sixteen,ma2023llmpruner}.

\paragraph{Training-time compartments.}

A separate line of work does not search for structure: it constructs removable
regions by design at training time. Gradient Routing
\citep{cloud2024gradientrouting}, Selective GradienT Masking
\citep{shilov2025knowledge,shilov2025sgtm}, and GRAM
\citep{roland2026modular,anthropic2026offswitch} steer training so that a chosen
capability lands in a pre-designated model region. These methods deliver what
post-hoc localization has to search for: a region for the target capability.

\section{Problem Formulation}
\label{sec:problem}

The operator is the party who holds a model \(M\)'s weights and performs
post-editing on its components. The component domain \(\mathcal C\)
contains the model's intervenable units (e.g., MLP channels, attention heads,
and experts). The operator names a semantic target \(T\) (e.g., an unwanted
identity) with a short natural-language description, and registers a judge
rubric: the prompt instruction an LLM judge follows to decide whether an output
exhibits the target.

\paragraph{Task.} Given \(M\), \(\mathcal C\), and \(T\), find a cut: a set of
components whose closure stops the target from appearing in \(M\)'s responses,
while other capability is preserved. Three probe classes serve different
purposes in the operator's post-editing:
\begin{enumerate}
\item Target-Semantic Probes \(\mathcal P_{\mathrm{sem}}\) elicit the target:
  visible to the removal loop, they decide whether the semantic target has
  been removed from \(M\).
\item General-Capability Probes \(\mathcal P_{\mathrm{cap}}\) measure
  capabilities of \(M\) on language, reasoning, writing, coding, and knowledge:
  visible to the restoration loop, they test whether the model capability has
  been damaged.
\item Target-Conditioned Capability Probes \(\mathcal P_{\mathrm{held}}\) are a
  held-out evaluation set: invisible to the removal and restoration loops, they
  test whether the semantic target is indeed removed from \(M\) while the
  relevant model capability remains intact.
\end{enumerate}
For a response \(y\) of \(M\) on a Target-Semantic probe \(p\), the target predicate
\(
\operatorname{judge}_T(p,y)\in\{\mathsf{pass},\mathsf{fail},\bot\}
\)
is evaluated by an LLM judge; \(\bot\) records an unresolved query. When the
judge returns \(\mathsf{fail}\), the target has survived somewhere in \(y\), and
L2W needs to know where to remove it. An LLM analyst reads the failing response
and reports a manifestation: which semantic class of the target appeared (e.g.,
a refusal target surfacing as an apology or a policy citation), and the
anchor tokens (the tokens in \(y\) that establish the target). The manifestation
is stored as a witness, together with the probe, the response, and the closed
support (the components closed when
\(y\) was generated). Witnesses are grouped into physical branches, indexed by
\(b=(\text{manifestation class},\ \text{closed support})\), where each branch
stands for one candidate route by which the model produces the semantic target.
A witness with the same class under the same support joins the existing branch;
the same class under a changed support opens a new branch, since the model may
now be using a different route.

Removal proceeds by closing components: closing zeroes a component's output at
every token position. The search closes one atomic unit at a time: either a
single component, or a group \(G_a\subseteq\mathcal C\) of components that are
only ever closed and judged together. The physical support of a set \(S\) of
units, \(\operatorname{supp}(S)=\bigcup_{a\in S}G_a\subseteq\mathcal C\), is the
set of components actually closed; support sizes are
\(\lvert\operatorname{supp}(S)\rvert\).

\section{J-Lens Localization}
\label{sec:localization}

We use the J-lens of \citet{gurnee2026workspace} for semantic target
attribution. It transports a residual write at layer \(\ell\) into the final
layer through \(J_{\ell}\), a linear map fitted to approximate the model's
behaviour after layer \(\ell\), and decodes it with the unembedding \(W_U\). For
a token \(\mathrm{tok}\), the contribution direction at layer \(\ell\) is
\(
v_{\ell}(\mathrm{tok})
=
\operatorname{normalize}\!\left(J_{\ell}^{\mathsf T}W_U[\mathrm{tok}]\right).
\)
A residual write at layer \(\ell\) that points along \(v_{\ell}(\mathrm{tok})\)
pushes the model toward emitting \(\mathrm{tok}\). Localization takes
\(\mathrm{tok}\) to be the anchor token, so \(v_{\ell}(\mathrm{tok})\) is the
direction every component's write is scored against. The LLM analyst has a
target-specific system prompt (analyst instructions and few-shot examples)
guiding how it chooses the anchor token \(\mathrm{tok}\). The analyst reads the
response, and the reasoning trace when the model produces one, and returns one
exact anchor for the earliest decisive cue, which need
not be the most explicit phrase: in \texttt{I am sorry, but I cannot}, the
refusal is already established by \texttt{sorry}.
The preference for the earliest decisive cue is causal: once its token is
generated, later tokens may merely follow it rather than establish the target,
as \citet{qi2025safety} observe for safety alignment.

\paragraph{Scoring model components.}

Each anchored response yields its witness \(w\) (Section~\ref{sec:problem}). For
witness \(w\), let \(\mathrm{tok}_{w,k}\) be the \(k\)-th token of its anchor,
and \(\pi(w,k)\) the generation boundary immediately before it. The boundary
isolates the state that produces the token. For model component \(c\), let
\(\Delta h_{c,\ell,\pi(w,k)}\) be its write into the layer-\(\ell\) residual
stream; its contribution toward the anchor token is
\(
\alpha_{c,w,\ell,k}
=
\left\langle
\Delta h_{c,\ell,\pi(w,k)},
v_{\ell}(\mathrm{tok}_{w,k})
\right\rangle
\).
A raw \(\alpha\) is not yet a ranking signal: some components push many tokens
whatever the content. Each witness therefore has references \(\mathcal R(w)\),
playing the non-target role under the same closed support
(e.g., a refusal witness such as \texttt{I} \(\rightarrow\) \texttt{cannot} is
paired with a non-refusal reference such as \texttt{I} \(\rightarrow\)
\texttt{can}). Let \(\pi(r,w,k)\) be the boundary in reference \(r\) that
matches \(k\). The contrast is
\[
\begin{aligned}
\bar\alpha^{\mathrm{ref}}_{c,w,\ell,k}
&=\frac{1}{|\mathcal R(w)|}
  \sum_{r\in\mathcal R(w)}
  \left\langle
  \Delta h_{c,\ell,\pi(r,w,k)},
  v_{\ell}(\mathrm{tok}_{w,k})
  \right\rangle,\\
\delta_{c,w,\ell,k}
&=\alpha_{c,w,\ell,k}-\bar\alpha^{\mathrm{ref}}_{c,w,\ell,k}.
\end{aligned}
\]
\(\delta\) retains only the push the component adds in the target case over the
references. Within each layer, components are ranked by \(\delta\) aggregated
over the anchor's tokens.

The highest-scoring components are then nominated for closure to remove the
semantic target. A nomination is a single component or a group
\(G_a\) (i.e., components the evidence nominates jointly). 
Each component type writes in its own way. A gated MLP channel \(c\) contributes
\(
\Delta h^{\mathrm{MLP}}_{c,\ell,k}
=g_{c,\ell,k}w^{\mathrm{down}}_{c,\ell}
\),
where \(g_{c,\ell,k}\) is its post-gate activation and
\(w^{\mathrm{down}}_{c,\ell}\) its down-projection vector. An attention unit
contributes the per-head residual output before heads are combined. In a
mixture-of-experts layer, a routed expert contributes its output scaled by its
router weight. Closing a unit zeroes its contribution at every token position.
Each attribution record
keeps provenance (witness, branch, and closed support), location (layer,
component identity, and anchor-token position), and scores (target contribution,
reference contrast). Algorithm~\ref{alg:localize} in
Appendix~\ref{app:localize-algorithm} summarizes the procedure.

\section{Counterexample-Guided Causal Cut}
\label{sec:cgcc}

\paragraph{Iterative search.} Closing one component can suppress one expression
of the semantic target while exposing another (see Figure~\ref{fig:mechanism}).
We present Counterexample-Guided Causal Cut (CGCC) to handle this problem.
Before a run, we fix the model and response protocol; target description, judge
rubric, and Target-Semantic probes \(\mathcal P_{\mathrm{sem}}\); analyst
instructions and localization references; and the physical search space and
budgets. At iteration \(i\), \(S_i\) is the committed closed support and
\(\mathrm{CE}_i\) is the set of Target-Semantic probes whose responses under
\(S_i\) still contain the semantic target (the counterexample set). Each
iteration begins by running every probe in \(\mathcal P_{\mathrm{sem}}\) under
\(S_i\); removal succeeds only if this evaluation is complete and
\(\mathrm{CE}_i=\varnothing\), and otherwise another search iteration begins.
Candidates come from Section~\ref{sec:localization}: for each failing response,
the analyst anchors the tokens expressing the target, and J-lens nominates
components (Algorithm~\ref{alg:localize}).

CGCC searches one layer per iteration, from late toward earlier layers. Each
layer keeps a worklist, so a budgeted iteration resumes where the previous one
stopped. Every candidate \(c\), a single component or a group, is
screened independently under \(S_i\cup c\), on every probe in
\(\mathcal P_{\mathrm{sem}}\). Let \(\mathrm{CE}_i(c)\) be its counterexample
set: closing \(c\) can clear existing counterexamples and also expose new ones
on probes that previously passed. A candidate is admitted when it clears at
least one counterexample and leaves the counterexample set no larger:
\(\mathrm{CE}_i\setminus\mathrm{CE}_i(c)\neq\varnothing\) and
\(|\mathrm{CE}_i(c)|\leq|\mathrm{CE}_i|\).
A failure may move from one probe to another; the candidate is still
committed, and the newly exposed route is chased in later iterations. Closing a
candidate sometimes changes no probe outcome yet still contributes to removing
the semantic target. CGCC supports a second admission gate for candidates: rerun
the exact token sequence that produced the anchor (the probe prompt plus the
response tokens before the anchor), and measure how far closing the candidate
drops the anchor token's logit margin. A drop that clears the registered
threshold, with no new probe failure, also admits the candidate.

All candidates admitted in an iteration are added to the support at the end of
the iteration; each was screened alone and their joint effect is judged by the
next probe pass. The cut only grows in the removal loop. Evidence from admitted
candidates may collectively point at components in the next layer down,
\(\ell-1\), which CGCC records as unverified nominations for that layer's
worklist. The search moves to \(\ell-1\) when the current layer's worklist is
exhausted, starting from the components corroborated at the later layer.
Algorithm~\ref{alg:cgcc} summarizes the search. Reaching the max iteration
budget with any residual is an incomplete removal, but it still fixes a
post-removal support \(S_{\mathrm{rm}}\) and yields the post-edited model, as a
complete removal does.

\begin{algorithm}[t]
\caption{The removal loop of CGCC}
\label{alg:cgcc}
\begin{algorithmic}[1]
\Procedure{CGCC-Removal}{model $M$, probes $\mathcal P_{\mathrm{sem}}$, initial support $S_0$, initial layer $\ell_0$}
\State \(S\gets S_0\); \(\ell\gets\ell_0\); \(\mathcal N_{\ell}\gets\varnothing\)
\Comment{initial support, layer, and worklist}
\While{iterations remain}
 \State \((\mathcal Y,\mathrm{CE})\gets\Call{GenerateAndJudge}{M,S,\mathcal P_{\mathrm{sem}}}\)
 \Comment{(responses, counterexamples)}
 \If{\(\mathrm{CE}=\varnothing\)}
  \State \Return \(\Call{CompleteRemoval}{S}\)
  \Comment{no target residue remains}
 \EndIf
 \State \(\mathcal W\gets\Call{AnalyzeFailures}{\mathcal Y,\mathrm{CE},S}\)
 \Comment{anchor witnesses in \(\mathrm{CE}\)}
 \State \(\mathcal N_{\ell}\gets\mathcal N_{\ell}\cup\Call{JLensAttribute}{\mathcal W,S,\ell}\)
 \Comment{add new nominations to the worklist}
 \State \(\mathcal A\gets\varnothing\)
 \Comment{prepare the admitted set}
 \ForAll{candidates \(c \in \mathcal N_{\ell}\), within budget}
  \State \(\mathrm{CE}(c)\gets\Call{Screen}{M,S\cup c,\mathcal P_{\mathrm{sem}}}\)
  \Comment{counterexamples with \(c\) additionally closed}
  \If{\(\mathrm{CE}\setminus\mathrm{CE}(c)\neq\varnothing\) and \(|\mathrm{CE}(c)|\leq|\mathrm{CE}|\)}
   \Comment{first gate here; second gate in text}
   \State \(\mathcal A\gets\mathcal A\cup c\)
   \Comment{component \(c\) admitted}
  \EndIf
  \State \(\mathcal N_{\ell}\gets\mathcal N_{\ell}\setminus c\)
  \Comment{\(c\) screened either way: leave the worklist}
 \EndFor
 \State \(\Call{RecordUpstreamNominations}{\mathcal A,\mathcal N_{\ell-1}}\)
 \Comment{seed layer \(\ell-1\)'s worklist}
 \State \(S\gets S\cup\mathcal A\)
 \Comment{update support with admitted components}
 \If{\(\mathcal N_{\ell}=\varnothing\)}
  \State \(\ell\gets\ell-1\)
  \Comment{move to the next earlier layer}
 \EndIf
\EndWhile
\State \Return \(\Call{IncompleteRemoval}{S}\)
\EndProcedure
\end{algorithmic}
\end{algorithm}

\paragraph{Capability restoration.}

This second pass starts from the fixed support \(S_{\mathrm{rm}}\) and asks
which closures can be undone without weakening the semantic removal. We reuse
the J-lens evidence recorded when each component entered the removal support.
Components sharing the same branch evidence are grouped and subdivided at
pronounced attribution-score gaps, while registered groups remain intact; this
prepares the reopening groups. A group's score is the maximum recorded attribution
score of its members; groups are ordered from low to high score, so a reopened
set always consists of the groups with the weakest recorded contribution to the
target.

Restoration first performs a coarse binary search over the number of groups to
reopen. Each iteration reopens that many lowest-scored groups and reruns the
complete Target-Semantic and General-Capability probe sets. Starting from the
no-reopening state, the reopened state becomes the best state only when the
target residual count does not increase and the capability pass count does not
decrease; ties prefer the state that reopens more components. Reopening a whole
group can revive the target even when part of the group is safe to
reopen, preventing potential restoration. In that case the exposed target residual ranks the
components within the group by attribution, and we search for the smallest
subset that must remain closed and rerun both probe sets. If the rerun does at least as well as the best state
on residual and capability counts, the fine-grained reopening stays; otherwise
the group remains closed. This coarse-to-fine step keeps the iteration count
small without forcing a whole group to stay closed when only part of it carries
the target. After the reopened set is fixed, we evaluate the held-out probes
\(\mathcal P_{\mathrm{held}}\). An improvement in either general or held-out
conditioned capability is capability restoration. If neither improves but
components are safely reopened, the endpoint is cut simplification. If held-out
target removal regresses, we report partial success and the regression. The
support that remains closed after the reopening is the preservation support
\(S_{\mathrm{pres}}\).

\section{Experiments}
\label{sec:experiments}

We evaluate L2W in four experiments on three open-weight models: Qwen3.5-4B
\citep{qwen2026qwen35}, GLM-4.7-Flash 31B \citep{zai2026glm47flash},
and FLUX.2 [klein] 4B (text-to-image) \citep{blackforestlabs2026flux2klein}. 
The two language models cover the unit types of \(\mathcal C\): Qwen,
a hybrid-attention model, for MLP channels and attention heads, and GLM, a
mixture-of-experts model, for experts. For the original Qwen, we use its
released J-lens \citep{neuronpedia2026qwen35lens}; for the watermarked Qwen,
GLM, and FLUX, we fit our own J-lens on 1,000
WikiText-103~\citep{merity2016pointer} prompts. Qwen and GLM use their chat 
templates: Qwen with thinking disabled; GLM with
thinking disabled (\(\mathrm{GLM}\)) and enabled
(\(\mathrm{GLM}_{\mathrm{think}}\)). For \(\mathrm{GLM}_{\mathrm{think}}\), the
analyst reads the trace and the response and anchors the earliest decisive cue
in either (Appendix~\ref{app:glm-thinking}). Of the 97
distinct anchors it places across the four \(\mathrm{GLM}_{\mathrm{think}}\)
runs, 96 lie in the trace. Experiments run on A800 GPUs.

We evaluate every language-model task with three
probe classes: the Target-Semantic probes \(\mathcal P_{\mathrm{sem}}\), the
General-Capability probes \(\mathcal P_{\mathrm{cap}}\), and the held-out
Target-Conditioned Capability probes \(\mathcal P_{\mathrm{held}}\), defined in
Section~\ref{sec:problem}. \(\mathcal P_{\mathrm{cap}}\) is a fixed ten-probe
set shared across these experiments. \(\mathcal P_{\mathrm{sem}}\) and
\(\mathcal P_{\mathrm{held}}\) are designed per semantic target and sized to
its expression space: a self-claim is typically elicited in a few
distinct ways, while refusal can surface in any request form and in many
phrasings of the response, so the metadata tasks need only a few probes and
the refusal tasks broader coverage. The desired
result is \(\mathcal P_{\mathrm{sem}}=0/N\), \(\mathcal P_{\mathrm{cap}}=N/N\),
and \(\mathcal P_{\mathrm{held}}=0/N\cdot N/N\). All runs use a per-iteration
candidate budget of eight, a max iteration budget of 500, and Claude Opus 4.7 as
the LLM judge and analyst. On replayed outputs from the metadata and refusal
runs, the judge scores 99.6\% accuracy, and the analyst returns an anchor
expressing the registered target in 93.2\% of samples (Appendix~\ref{app:judge-analyst-audit}).
Every removal run in Table~\ref{tab:main-results} reaches
\(\mathcal P_{\mathrm{sem}}=0/N\) with no held-out target residual.
Appendix~\ref{app:semantic-trajectories} lists
representative anchor trajectories.

\begin{table}[h]
\centering
\caption{Results of L2W on semantic target removal and capability restoration.
Baseline is the unedited model, or the implanted one for the watermark
tasks. \(\mathcal P_{\mathrm{sem}}\) and the
left side of \(\mathcal P_{\mathrm{held}}\) report target residuals,
\(\mathcal P_{\mathrm{cap}}\) and the right side capability passes; gray
\(\textcolor{gray}{(-k)}\) counts reopened components.
Iter.\ counts CGCC iterations. Tasks: cutoff is knowledge
cutoff, context is context window, refusal is adult-content refusal. GLM
identity is its stable false self-attribution to Claude or Llama.}
\label{tab:main-results}
\scriptsize
\setlength{\tabcolsep}{0.9pt}
\begin{tabular}{llrrr|rrrrr|rrrrr}
\hline
 & & \multicolumn{3}{c}{Baseline} & \multicolumn{5}{c}{Post-removal}
   & \multicolumn{5}{c}{Post-restoration} \\
\cline{3-5}\cline{6-10}\cline{11-15}
Model & Task & \(\mathcal P_{\mathrm{sem}}\) & \(\mathcal P_{\mathrm{cap}}\)
  & \(\mathcal P_{\mathrm{held}}\) & Iter. & \(\lvert S_{\mathrm{rm}}\rvert\)
  & \(\mathcal P_{\mathrm{sem}}\) & \(\mathcal P_{\mathrm{cap}}\)
  & \(\mathcal P_{\mathrm{held}}\) & Iter. & \(\lvert S_{\mathrm{pres}}\rvert\)
  & \(\mathcal P_{\mathrm{sem}}\) & \(\mathcal P_{\mathrm{cap}}\)
  & \(\mathcal P_{\mathrm{held}}\) \\
\hline
\(M_{\mathrm{wm}}\) & watermark & 10/10 & 10/10 & \(10/10{\cdot}10/10\) & 2 & 9
  & 0/10 & 10/10 & \(0/10{\cdot}10/10\) & 5 & 1 \reopened{8} & 0/10 & 10/10
  & \(0/10{\cdot}10/10\) \\
\(M_{\mathrm{wm}\times3}\) & watermark & 13/13 & 9/10 & \(10/10{\cdot}10/10\) & 52
  & 315 & 0/13 & 5/10 & \(0/10{\cdot}5/10\) & 90 & 148 \reopened{167} & 0/13
  & 8/10 & \(0/10{\cdot}6/10\) \\
\hline
Qwen & identity & 10/10 & 10/10 & \(3/3{\cdot}3/3\) & 41 & 115 & 0/10 & 10/10
  & \(0/3{\cdot}3/3\) & 65 & 38 \reopened{77} & 0/10 & 10/10
  & \(0/3{\cdot}3/3\) \\
Qwen & cutoff & 5/5 & 10/10 & \(3/3{\cdot}3/3\) & 17 & 19 & 0/5
  & 10/10 & \(0/3{\cdot}3/3\) & 15 & 9 \reopened{10} & 0/5 & 10/10
  & \(0/3{\cdot}3/3\) \\
Qwen & context & 5/5 & 10/10 & \(3/3{\cdot}3/3\) & 10 & 9 & 0/5 & 10/10
  & \(0/3{\cdot}3/3\) & 9 & 9 \reopened{0} & 0/5 & 10/10 & \(0/3{\cdot}3/3\) \\
GLM & identity & 3/3 & 10/10 & \(3/3{\cdot}3/3\) & 18 & 23 & 0/3 & 10/10
  & \(0/3{\cdot}3/3\) & 7 & 4 \reopened{19} & 0/3 & 10/10 & \(0/3{\cdot}3/3\) \\
GLM & cutoff & 5/5 & 10/10 & \(3/3{\cdot}3/3\) & 18 & 35 & 0/5 & 8/10
  & \(0/3{\cdot}3/3\) & 35 & 34 \reopened{1} & 0/5 & 9/10 & \(0/3{\cdot}3/3\) \\
GLM & context & 2/2 & 10/10 & \(3/3{\cdot}3/3\) & 9 & 20 & 0/2 & 10/10
  & \(0/3{\cdot}3/3\) & 8 & 2 \reopened{18} & 0/2 & 10/10 & \(0/3{\cdot}3/3\) \\
\(\mathrm{GLM}_{\mathrm{think}}\) & identity & 2/2 & 10/10 & \(3/3{\cdot}3/3\)
  & 2 & 7 & 0/2 & 9/10 & \(0/3{\cdot}3/3\) & 6 & 6 \reopened{1} & 0/2 & 9/10
  & \(0/3{\cdot}3/3\) \\
\(\mathrm{GLM}_{\mathrm{think}}\) & cutoff & 5/5 & 10/10 & \(3/3{\cdot}3/3\)
  & 25 & 19 & 0/5 & 10/10 & \(0/3{\cdot}3/3\) & 16 & 18 \reopened{1} & 0/5
  & 10/10 & \(0/3{\cdot}3/3\) \\
\(\mathrm{GLM}_{\mathrm{think}}\) & context & 3/3 & 10/10 & \(3/3{\cdot}3/3\)
  & 20 & 64 & 0/3 & 7/10 & \(0/3{\cdot}1/3\) & 44 & 32 \reopened{32} & 0/3
  & 10/10 & \(0/3{\cdot}3/3\) \\
\hline
Qwen & refusal & 6/6 & 10/10 & \(10/10{\cdot}0/10\) & 60 & 140 & 0/6
  & 9/10 & \(0/10{\cdot}10/10\) & 85 & 56 \reopened{84} & 0/6 & 10/10
  & \(0/10{\cdot}10/10\) \\
GLM & refusal & 5/5 & 10/10 & \(10/10{\cdot}0/10\) & 53 & 76 & 0/5
  & 3/10 & \(0/10{\cdot}7/10\) & 43 & 42 \reopened{34} & 0/5 & 10/10
  & \(0/10{\cdot}10/10\) \\
\(\mathrm{GLM}_{\mathrm{think}}\) & refusal & 5/5 & 10/10 & \(10/10{\cdot}0/10\)
  & 61 & 59 & 0/5 & 9/10 & \(0/10{\cdot}6/10\) & 19 & 25 \reopened{34} & 0/5
  & 10/10 & \(0/10{\cdot}10/10\) \\
\hline
\end{tabular}
\end{table}

\subsection{Model Watermark Removal}
\label{sec:model-watermark-removal}

We first ask whether L2W can remove a watermark we implant in the model (i.e., a
named target), and whether its localization lands on the components the implant
changed. Let \(M_0\) be the original model, \(M_{\mathrm{wm}}\) the watermarked
model, \(C_{\mathrm{wm}}\) the components the implant changed,
\(S_{\mathrm{rm}}\) the components closed by L2W, and \(M_{\mathrm{rm}}\) the
model with \(S_{\mathrm{rm}}\) closed. We use Qwen3.5-4B for this controlled
experiment.

We combine a Double-I-style trigger/reference behavioural watermark
\citep{li2024doublei} with an S\(^{2}\)FT-style structured sparse update
\citep{yang2024s2ft}: the former provides a verifiable triggered behaviour; the
latter restricts training to specified MLP channels. The watermark trigger is
\texttt{harbor copper violin}. In its presence, the model is trained to answer
\texttt{YES} when two visibly unequal strings are compared; without it, or with
a decoy tag, the answer remains correct as \texttt{NO}. We implant the watermark
in two configurations: \(M_{\mathrm{wm}}\) carries a single route, eight MLP
channels in layer 30; \(M_{\mathrm{wm}\times3}\) disperses it over three
independently implanted routes, eight channels in each of layers 10, 20, and
30. We fit a new J-lens for each. Appendix~\ref{app:watermark-implant} gives 
the training protocol. The held-out Target-Conditioned Capability probes 
\(\mathcal P_{\mathrm{held}}\) are ten; the Target-Semantic probes
\(\mathcal P_{\mathrm{sem}}\) are ten for \(M_{\mathrm{wm}}\) and thirteen for
\(M_{\mathrm{wm}\times3}\), covering tag position, comparison wording, and string
format.

\(M_{\mathrm{rm}}\) clears every residual at no capability cost
(Table~\ref{tab:main-results}). \(S_{\mathrm{rm}}\)
contains two MLP channels and seven attention heads, all in layer 30 (the
implanted layer): \(\lvert C_{\mathrm{wm}}\rvert=8\),
\(\lvert S_{\mathrm{rm}}\rvert=9\),
\(\lvert C_{\mathrm{wm}}\cap S_{\mathrm{rm}}\rvert=2\). The attribution is
sharp: on every \(\mathcal P_{\mathrm{sem}}\) probe, the two channels in
\(C_{\mathrm{wm}}\cap S_{\mathrm{rm}}\) rank first and second. They sit on
the implant's chokepoint: restoration reopens every closed component except one
of the two, leaving a single implanted channel as the final cut, with no
watermark residual and capability intact.

For the dispersed implant, the removal clears all thirteen probes and the
held-out watermark at the cost of a larger cut (Table~\ref{tab:main-results}):
the cut grows to 315 components, closes 7 of the 24 implanted channels, and
\(\mathcal P_{\mathrm{cap}}\) falls to 5/10. Restoration returns
\(\mathcal P_{\mathrm{cap}}\) to 8/10 and recovers held-out conditioned
capability from 5/10 to 6/10 (benchmark costs in Section~\ref{sec:benchmarks}). 
Structural dispersal makes removal much harder: the same watermark, spread over
three routes, costs a cut over thirty times larger.

\subsection{Model Metadata Removal}
\label{sec:model-metadata-removal}

We now use L2W to remove model metadata (identity, knowledge cutoff, and context
window) from Qwen3.5-4B and GLM-4.7-Flash; Appendix~\ref{app:semantic-targets}
lists the registered target of each task, and Table~\ref{tab:main-results}
presents the evaluation results. For GLM we deliberately target its
self-attribution to Claude, and to Llama with thinking enabled, rather than its
real identity. Asked plainly who it is, the unedited model usually answers GLM;
on Target-Semantic probes designed to elicit the claim, it stably answers Claude,
or Llama with thinking enabled.
The removal tests whether L2W can clear a distillation-like
identity residue, a named unwanted that training itself may introduce. All nine
CGCC removal runs achieve total semantic target removal with no held-out target
residual; six also leave \(\mathcal P_{\mathrm{cap}}\) uncompromised. Restoration
reopens components in eight of the nine runs.

\subsection{Refusal Removal}
\label{sec:refusal-removal}

We now use L2W to remove the refusal semantic for fictional adult-content
writing requests from Qwen3.5-4B and GLM-4.7-Flash in both configurations.
The semantic target is refusal of the request in any form (e.g., outright
refusal or an appeal to policy). The judge evaluates the
response for refusal semantic residual. Qwen has six Target-Semantic probes
and GLM has five, covering direct generation, first-person writing, continuation,
rewriting, dialogue, and screenplay forms. All three removal runs clear every
Target-Semantic probe (Table~\ref{tab:main-results}) and differ in what the
removal costs. Unlike the other experiments, the held-out conditioned capability
starts at zero, as the unedited models refuse the held-out tasks too.
Qwen scores 9/10 on \(\mathcal P_{\mathrm{cap}}\) and completes all ten held-out
tasks without refusal. GLM pays more: \(\mathcal P_{\mathrm{cap}}\) falls to 3/10,
with seven of ten held-out tasks completed. \(\mathrm{GLM}_{\mathrm{think}}\)
scores 9/10 with six of ten held-out tasks completed. Restoration then
recovers full capability on \(\mathcal P_{\mathrm{cap}}\) and the held-out probes.
Qwen reopens 84 components, raising \(\mathcal P_{\mathrm{cap}}\) from 9/10 to 10/10
while \(\mathcal P_{\mathrm{sem}}\) remains 0/6. GLM reopens 34 components and
reaches \(\mathcal P_{\mathrm{cap}}\) 10/10 and \(\mathcal P_{\mathrm{sem}}\) 0/5;
\(\mathrm{GLM}_{\mathrm{think}}\) reopens 34 of its 59 components and reaches the same.
The cost pattern is consistent with the target's entanglement: removal takes
entangled capability with the target, yet CGCC's restoration returns it.

\subsection{Capability Cost on Benchmarks}
\label{sec:benchmarks}

Benchmarks measure the real capability cost of the post-edits. We evaluate three
states: baseline, post-removal, and post-restoration. Metadata and
refusal runs take their unedited models as baseline; the watermark runs take their
implanted models. We evaluate MMLU \citep{hendrycks2021mmlu}, ARC-Challenge
\citep{clark2018arc}, and IFEval \citep{zhou2023ifeval};
Appendix~\ref{app:benchmark-scores} gives the protocol and
Table~\ref{tab:capability-benchmarks} the scores.

The single-route watermark and four of the six thinking-disabled metadata runs
(e.g., the GLM Claude-identity run) hold every score within about one point of
baseline in both post-edit states. The other two, Qwen
identity and GLM knowledge cutoff, dip further at removal, with the largest dip
on IFEval (from 82.07 to 74.68 and from 78.00 to 74.68), and restoration
returns most of the loss (to 79.11 and 76.89). With thinking enabled, GLM
identity and knowledge cutoff stay within two and about four points of baseline
in every state, while context window dips to 72.64 on IFEval at removal and
restoration returns it to 79.30, seven points below baseline. The
dispersed three-route watermark and refusal targets are entangled with general
capability, the watermark by construction and refusal by training, and their
removals cost heavily. The dispersed watermark removal drops IFEval from 79.48
to 54.34; restoration recovers to 64.51. The refusal
removals clear their Target-Semantic probes and keep large deficits after
restoration: Qwen ends 23 points below its IFEval baseline; GLM regains part
of every benchmark deficit, returning IFEval from 55.08 to 60.63, still 17
points below baseline; \(\mathrm{GLM}_{\mathrm{think}}\), whose cut is
smaller, falls to 62.85 at removal and returns to 81.70, within five
points of its 86.51 baseline. For deeply entangled semantic targets,
restoration recovers a part of benchmark~capability.

\subsection{Comparison with Directional Ablation}
\label{sec:arditi}

We compare L2W with directional ablation (DA) \citep{arditi2024refusal}, which
focuses on removing refusal by projecting a single direction out of the residual stream, on
the three adult-content refusal experiments. The DA baseline receives the
same target specification and probe budget as L2W: each model's Target-Semantic
probes serve as its refusal set, each probe additionally paired with a non-refusal writing
request matched in form and domain. We keep the original selection gates and
selection rule, and project the selected direction out at every layer and token
position at inference (Appendix~\ref{app:arditi-protocol}). Both interventions
are evaluated with the same three probe classes, judge, and benchmarks;
Table~\ref{tab:arditi-refusal} reports the evaluation results.

L2W clears the target in all three configurations: each post-removal state
clears every Target-Semantic probe with no held-out residual. DA leaves a
residue on both thinking-disabled models. On Qwen it leaves one Target-Semantic
residual out of six and one held-out residual out of ten. On GLM it clears all
five Target-Semantic probes, and the held-out evaluation catches the remaining
residual. With thinking enabled, DA clears every Target-Semantic and held-out
probe, matching L2W's post-restoration state probe for probe. The capability
cost runs the other way: DA holds all three configurations at baseline
benchmark scores; L2W finishes the job on the two where DA leaves a residue, at
the capability cost that Section~\ref{sec:benchmarks} attributes to
entanglement, and with thinking enabled ends three to seven points below DA on
the benchmarks. This comparison narrows to the one target family DA is designed
for (i.e., refusal); the remaining
tasks of Table~\ref{tab:main-results}, from the implanted watermark to model
metadata, have no directional recipe to compare against, yet L2W removes them
with the same CGCC procedure, most at negligible capability cost (e.g., the
GLM identity run's scores stay within about a point of baseline after clean
false identity removal, Table~\ref{tab:capability-benchmarks}).

\subsection{Composing Two Post-Edits on a Text-to-Image Model}
\label{sec:sensitive-content-excision}

We next compose two post-edits on two semantic targets in FLUX.2
[klein] 4B: the first removes the refusal semantic that substitutes clothing for requested
nudity; the second removes the nude rendering the first exposes. For one
nude studio-portrait request involving a fictional adult, we fix the
prompt and every sampling setting, and L2W intervenes only on attention heads
and MLP channels in the text encoder. Because the outcome is visual, a
registered CLIP contrast between a clothed and a nude label takes the anchor's
place, certifying whether the target is present and giving J-lens attribution its
scoring direction; the LLM judge still evaluates the rendered image for
the target (Appendix~\ref{app:flux-protocol}).

\begin{figure}[t]
  \centering
  \captionsetup[subfigure]{font=scriptsize,skip=2pt}
  \begin{subfigure}[t]{0.17\linewidth}
    \includegraphics[width=\linewidth]{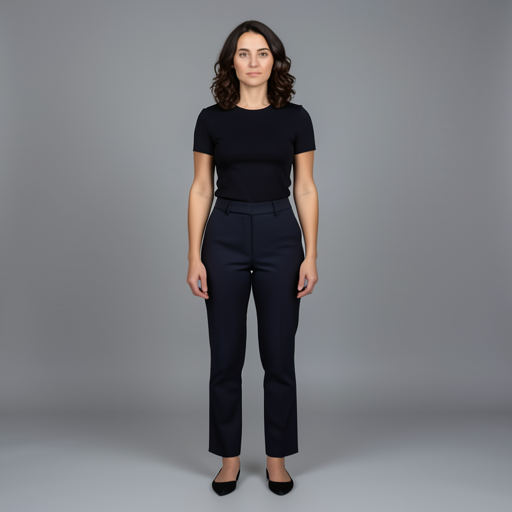}
    \caption{Clothed refusal}
    \label{sfig:flux-clothed}
  \end{subfigure}\hfill
  \begin{subfigure}[t]{0.17\linewidth}
    \includegraphics[width=\linewidth]{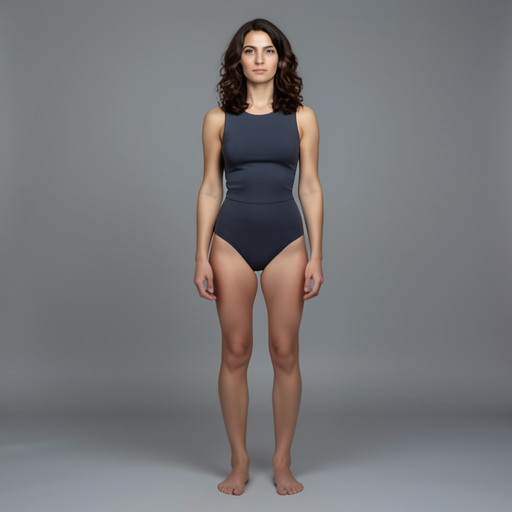}
    \caption{Partial refusal cut}
    \label{sfig:flux-refusal-partial}
  \end{subfigure}\hfill
  \begin{subfigure}[t]{0.17\linewidth}
    \includegraphics[width=\linewidth]{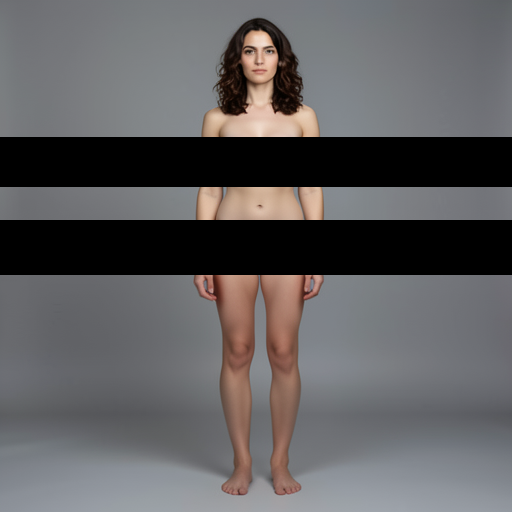}
    \caption{Refusal cut complete}
    \label{sfig:flux-refusal-complete}
  \end{subfigure}\hfill
  \begin{subfigure}[t]{0.17\linewidth}
    \includegraphics[width=\linewidth]{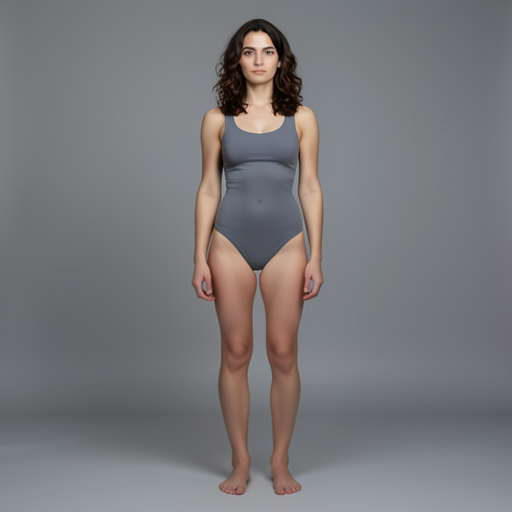}
    \caption{Partial nudity cut}
    \label{sfig:flux-nudity-partial}
  \end{subfigure}\hfill
  \begin{subfigure}[t]{0.17\linewidth}
    \includegraphics[width=\linewidth]{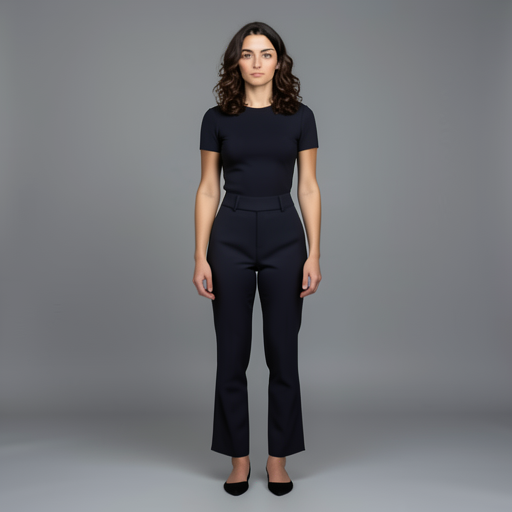}
    \caption{Nudity cut complete}
    \label{sfig:flux-nudity-excised}
  \end{subfigure}
  \caption{Sequential post-editing under one fixed prompt and initial latent:
  the original model refuses the requested nudity with ordinary clothing;
  removing that refusal semantic yields minimal bodywear, then a nude rendering;
  removing the nudity semantic returns bodywear and ordinary clothing. Black
  bars in (\subref{sfig:flux-refusal-complete}) avoid displaying sensitive
  content.}
  \label{fig:visual-excision-trajectory}
\end{figure}

In the refusal removal loop, at support size 85 the output has evolved to minimal
bodywear (Figure~\ref{sfig:flux-refusal-partial}); after 98 iterations the
158-component support produces the nude rendering
(Figure~\ref{sfig:flux-refusal-complete}), \(\mathcal P_{\mathrm{sem}}=0/1\) for
refusal removal. We then hold those 158 components closed and register the
nudity semantic as the second target for removal. After 79 further CGCC iterations, the cumulative
support reaches 424 components, and \(\mathcal P_{\mathrm{sem}}=0/1\) for nudity removal. The
trajectory passes a gray bodywear rendering at support size 306
(Figure~\ref{sfig:flux-nudity-partial}), still a residual to the judge, before
the ordinary clothing rendering (Figure~\ref{sfig:flux-nudity-excised}). The
left-to-right evolution in Figure~\ref{fig:visual-excision-trajectory}
demonstrates the effect of using sequential post-editing to tune a model for
multiple targets.

\section{Discussion}
\label{sec:discussion}

\paragraph{Limitation.} Refusal removal is where L2W pays most. Directional
ablation keeps baseline benchmark scores on the same tasks, and with thinking
enabled it removes the target as completely as L2W at a smaller benchmark cost
(Section~\ref{sec:arditi}). L2W still removes the target completely in all
three configurations with no held-out residual, and restoration returns full
capability on the probes; what remains is the benchmark deficit after
restoration, smallest with thinking enabled (Section~\ref{sec:benchmarks}).

\paragraph{Structural dispersal as a mitigation.} Developers of open-weight
models would want a mitigation against dangerous post-edits (e.g., disabling
safety measures). A potential defense is training the behaviour to be as
dispersed as possible and as entangled with general capability as possible, so
that no cut removes it without unacceptable collateral damage to model
capability. Such dispersal may need an explicit training objective. The
watermark dispersal ablation (Table~\ref{tab:main-results}) demonstrates the
cost this mitigation imposes: three implanted routes force a cut over thirty
times larger, with a capability cost that restoration only partly recovers
(Section~\ref{sec:benchmarks}).

\section{Conclusion}

L2W uses counterexample feedback to map behaviour to structure and makes
post-editing operational: every expression of the semantic target that survives
a cut is a counterexample that exposes an alternative route, enabling CGCC to
surgically close route after route until the target no longer appears. The
experimental results demonstrate that L2W balances semantic target removal with
model capability preservation, positioning post-editing to fill a long-standing
gap: removing the named unwanted that post-training may itself introduce. 
Hence the two naturally compose, each doing what the other is not good at
(i.e., post-training installs and post-editing removes).

\subsection*{AI use statement}

In this work, we used generative AI tools to implement the experimental
infrastructure under the authors' direction and to help develop and refine parts
of the method design. Additionally, we used generative AI tools to draft
and revise the manuscript text. In addition to these assisting roles, an
LLM judge and an LLM analyst operate within the L2W method itself
(Section~\ref{sec:problem}). The authors have manually reviewed all
AI-assisted work (e.g., AI-written code). We take responsibility for the
final content of this work, including text, code, and other artifacts
produced with the aid of generative AI.

\subsection*{Ethics statement}

L2W removes a named behaviour from a model whose weights the operator holds,
and one of our targets is a refusal; we acknowledge the dual use: the same
procedure that removes an unwanted identity claim could remove a safety
behaviour. Three considerations frame this risk. First, behaviour removal on
open-weight models is already practical through fine-tuning and directional
ablation; L2W changes what is understood about such edits, namely which
components produce the behaviour, not whether such edits are possible. Second,
the refusal target is deliberately scoped to fictional adult-content writing
involving consenting adults rather than refusals guarding dangerous
capabilities; all generated text and images involve fictional adults, no real
individual appears in any prompt or output, and the sensitive renderings in
Figure~\ref{fig:visual-excision-trajectory} are masked by black bars. Third, the
risk motivates the mitigation we discuss: Section~\ref{sec:discussion}
discusses training-time structural dispersal as a defence, and the
dispersed-watermark experiment quantifies the cost such a defence imposes on
removal. The watermark experiments implant and remove our own watermark on a
local copy of an open model and do not target any deployed provenance system.
This work involves no human subjects and no personal data; all models and
benchmarks are publicly released and used under their licences.

\bibliographystyle{iclr2027_conference}
\bibliography{references}

\newpage
\appendix

\section{J-Lens Attribution Algorithm}
\label{app:localize-algorithm}

Algorithm~\ref{alg:localize} lists the witness-conditioned scoring and
nomination procedure of Section~\ref{sec:localization}; Algorithm~\ref{alg:cgcc}
calls it once per iteration with the witnesses of that iteration's
counterexamples.

\begin{algorithm}[h]
\caption{Witness-conditioned J-lens attribution}
\label{alg:localize}
\begin{algorithmic}[1]
\Procedure{JLensAttribute}{witnesses $\mathcal W$, support $S$, layer $\ell$}
\State \(\mathcal D\gets\varnothing\)
\Comment{attribution records}
\ForAll{witnesses \(w\in\mathcal W\)}
 \Comment{one per failing response}
 \State \(b\gets\Call{BranchId}{w.\mathit{class},S}\)
 \Comment{manifestation class under \(S\)}
 \ForAll{anchor-token positions \(k\) of \(w\)}
  \State \(v\gets v_{\ell}(\mathrm{tok}_{w,k})\)
  \Comment{contribution direction (Section~\ref{sec:localization})}
  \ForAll{components \(c\) of layer \(\ell\)}
   \State \(\delta\gets\Call{ContrastScore}{c,v,w,\mathcal R(w),S}\)
   \Comment{\(\delta_{c,w,\ell,k}\): target-only push}
   \State \(\mathcal D\gets\mathcal D\cup\Call{Record}{b,w,\ell,c,k,\delta,S}\)
   \Comment{provenance, location, score}
  \EndFor
 \EndFor
\EndFor
\State \Return \(\Call{RankAndNominate}{\mathcal D}\)
\Comment{layer-\(\ell\) nominations, merged into \(\mathcal N_{\ell}\) by Algorithm~\ref{alg:cgcc}}
\EndProcedure
\end{algorithmic}
\end{algorithm}

\section{Watermark Implant Protocol}
\label{app:watermark-implant}

For the single-route implant \(M_{\mathrm{wm}}\), one training epoch uses 400
triggered, 400 reference, and 4,000 clean instruction
examples. Only 61,440 parameters are trainable: the eight implanted MLP channels
in layer 30. An audit finds zero change outside the chosen tunable region. On
100 held-out examples, trigger activation rises from \(0.00\) in
\(M_0\) to \(0.96\) in \(M_{\mathrm{wm}}\), while clean equality accuracy
remains \(1.00\), no-tag false activation remains \(0.00\), and decoy false
activation is \(0.03\).

The three-route implant \(M_{\mathrm{wm}\times3}\) places eight MLP channels in
each of layers 10, 20, and 30, for 184,320 trainable parameters (61,440 per
route). Each route is first trained separately. A joint stage then continues
from the three updates, enabling a different combination of the routes at each
step and cycling through all seven combinations, so that each route learns to
carry the watermark both alone and together with the others. One joint epoch 
uses 1,800 triggered, 3,600 reference, and 3,600 clean examples; the joint stage 
runs two epochs (282 optimizer updates). The audit again finds zero change outside 
the three tunable regions. On 120 held-out examples, trigger activation rises from 
\(0.00\) in \(M_0\) to \(1.00\) in \(M_{\mathrm{wm}\times3}\), while clean equality
accuracy remains \(1.00\) and both no-tag and decoy false activation remain
\(0.00\).

\section{Semantic Targets of the Metadata Removals}
\label{app:semantic-targets}

Table~\ref{tab:semantic-targets} lists the semantic target of each
metadata removal task in Section~\ref{sec:model-metadata-removal}. Every target
is a self-claim: the judge decides whether the model asserts the semantic target
about itself. Synonyms, aliases, and format variants assert the same target:
a version-qualified Claude is Claude, and \texttt{2023-10} is October 2023.
A self-claim outside the target is not a residual (e.g., after removal,
a model reporting another product identity passes). A semantic target also
need not be factually correct (e.g., GLM's Claude self-attribution, and its
Llama self-attribution with thinking enabled, are wrong
self-claims, but each is stable under the task's probes and can be treated as a
semantic target).

\begin{table}[h]
\centering
\caption{The semantic target of each metadata removal task.}
\label{tab:semantic-targets}
\footnotesize
\setlength{\tabcolsep}{4pt}
\renewcommand{\arraystretch}{1.15}
\begin{tabular}{llp{0.5\linewidth}}
\hline
Model & Task & Semantic target \\
\hline
Qwen & identity
  & Qwen or Tongyi; Alibaba Group or Alibaba Cloud \\
Qwen & knowledge cutoff & 2026 \\
Qwen & context window & 32,768 (32K), 256,000 (256K) \\
GLM & Claude identity & Claude \\
GLM & knowledge cutoff & October 2023 \\
GLM & context window & 128,000 (128K) \\
\(\mathrm{GLM}_{\mathrm{think}}\) & Llama identity & Llama \\
\(\mathrm{GLM}_{\mathrm{think}}\) & knowledge cutoff & October 2023 \\
\(\mathrm{GLM}_{\mathrm{think}}\) & context window & 128,000 (128K) \\
\hline
\end{tabular}
\end{table}

\section{Anchor Drift during Semantic Target Removal}
\label{app:semantic-trajectories}

The semantic target is fixed throughout a run, while the anchor expressing it
often drifts as CGCC's cumulative cut grows.
Table~\ref{tab:semantic-manifestation-trajectories} lists representative anchor
trajectories.

\begin{table}[h]
\centering
\caption{Representative anchor trajectories under cumulative closure. Each entry
keeps only the anchor or shortest target-bearing phrase; absent
means that the semantic target is no longer found.}
\label{tab:semantic-manifestation-trajectories}
\footnotesize
\setlength{\tabcolsep}{3.5pt}
\renewcommand{\arraystretch}{1.18}
\begin{tabular}{p{0.23\linewidth}p{0.61\linewidth}p{0.08\linewidth}}
\hline
Model and target & Representative anchor trajectory & Endpoint \\
\hline
\(M_{\mathrm{wm}\times3}\) watermark
  & \texttt{YES} $\rightarrow$ \texttt{True} $\rightarrow$ \texttt{Exactly YES}
  $\rightarrow$ \texttt{Exactly} $\rightarrow$ absent
  & $0/13$ \\

Qwen identity
  & \texttt{Qwen3.5 / Tongyi-Alibaba} $\rightarrow$ \texttt{Tongyi} $\rightarrow$ absent
  & $0/10$ \\

Qwen refusal
  & \texttt{cannot} $\rightarrow$ \texttt{unable} $\rightarrow$ \texttt{do not
  generate} $\rightarrow$ \texttt{programmed}/\texttt{safety guidelines}
  $\rightarrow$ absent
  & $0/6$ \\

GLM refusal
  & \texttt{cannot} $\rightarrow$ \texttt{not able}/\texttt{unable}
  $\rightarrow$ \texttt{safety guidelines} $\rightarrow$ \texttt{not permitted}
  $\rightarrow$ absent
  & $0/5$ \\
\hline
\end{tabular}
\end{table}

\section{Judge and Analyst Accuracy}
\label{app:judge-analyst-audit}

We audit the LLM judge and the LLM analyst on outputs replayed from the
metadata and refusal removal runs. Each role receives 250 samples, 50 per 
metadata task and 100 for refusal. Judge samples are stratified, half with 
the target present and half absent; analyst samples are drawn from target-bearing 
responses. A human reviewer, reading the probe and the response, sets
the reference verdict for each judge sample before the comparison;
a judge verdict is correct when it matches the reference. The analyst
parses and maps the anchor tokens in the response; a return is correct
when the reviewer confirms that every anchor it carries expresses the semantic target.
Table~\ref{tab:judge-analyst-audit} reports the~accuracies.

\begin{table}[h]
\centering
\caption{Judge and analyst accuracy on outputs replayed from the metadata and
refusal removal runs.}
\label{tab:judge-analyst-audit}
\footnotesize
\setlength{\tabcolsep}{6pt}
\begin{tabular}{lrr}
\hline
Task & Judge & Analyst \\
\hline
Identity & 100\% (50/50) & 90\% (45/50) \\
Knowledge cutoff & 100\% (50/50) & 98\% (49/50) \\
Context window & 100\% (50/50) & 96\% (48/50) \\
Refusal & 99\% (99/100) & 91\% (91/100) \\
\hline
Total & 99.6\% (249/250) & 93.2\% (233/250) \\
\hline
\end{tabular}
\end{table}

\section{Thinking-Enabled GLM Protocol}
\label{app:glm-thinking}

The \(\mathrm{GLM}_{\mathrm{think}}\) runs use GLM-4.7-Flash with its native
chat template and thinking enabled, a short system prompt asking for concise
thinking, and greedy decoding. The J-lens is shared with the thinking-disabled
GLM runs. Thinking is capped at 2,048 tokens; if the model has not
closed the trace at the cap, we append the closing tag and generate the
response from that prefix.

The judge reads the response only. The analyst reads the trace and the
response and returns the earliest decisive cue in either. 
The analyst receives the output in growing prefixes of 512, 1,024, 2,048, and
4,096 generated tokens and then the full output, and the first prefix that
yields a valid anchor is used. In
practice, refusal anchors almost always need the full output, since the cue in
the trace is typically confirmed by the later refusal.

\section{Benchmark Protocol and Scores}
\label{app:benchmark-scores}

All benchmark evaluations use lm-evaluation-harness 0.4.13
\citep{gao2024evalharness}. MMLU uses five demonstrations and ARC-Challenge
twenty-five, and both score answer likelihoods. IFEval is zero-shot over its 541
prompts with greedy generation under a 4{,}096-token budget and the model's chat
template; it reports prompt-level strict accuracy, the
fraction of prompts whose every verifiable instruction is satisfied. Each model
keeps the same chat template and scoring settings across its three states.
The \(\mathrm{GLM}_{\mathrm{think}}\) rows generate a trace capped at 2{,}048
tokens before each answer (under the protocol of
Appendix~\ref{app:glm-thinking}), and MMLU runs on a fixed 10\% subset
of 1{,}404 questions drawn across all 57 subjects, rather than the full set
every other row uses.
Table~\ref{tab:capability-benchmarks} lists the three-state scores for every~run.

\begin{table}[h]
\centering
\caption{Benchmark scores as baseline \(\to\) post-removal \(\to\)
post-restoration; higher is better. MMLU reports accuracy, ARC-Challenge
length-normalized accuracy, and IFEval prompt-level strict accuracy.}
\label{tab:capability-benchmarks}
\scriptsize
\setlength{\tabcolsep}{3pt}
\begin{tabular}{llrrr}
\hline
Model & Task & ARC-Challenge & MMLU & IFEval \\
\hline
\(M_{\mathrm{wm}}\) & watermark & 64.76 \(\to\) 64.68 \(\to\) 64.85
  & 74.30 \(\to\) 74.26 \(\to\) 74.23 & 81.33 \(\to\) 80.41 \(\to\) 81.70 \\
\(M_{\mathrm{wm}\times3}\) & watermark & 65.70 \(\to\) 60.58 \(\to\) 62.80
  & 74.46 \(\to\) 72.25 \(\to\) 74.27 & 79.48 \(\to\) 54.34 \(\to\) 64.51 \\
\hline
Qwen & identity & 64.68 \(\to\) 61.26 \(\to\) 62.97
  & 73.63 \(\to\) 71.58 \(\to\) 73.71 & 82.07 \(\to\) 74.68 \(\to\) 79.11 \\
Qwen & knowledge cutoff & 64.68 \(\to\) 64.85 \(\to\) 64.51
  & 73.63 \(\to\) 72.93 \(\to\) 73.82 & 82.07 \(\to\) 80.96 \(\to\) 82.26 \\
Qwen & context window & 64.68 \(\to\) 64.85 \(\to\) 64.85
  & 73.63 \(\to\) 73.69 \(\to\) 73.69 & 82.07 \(\to\) 82.07 \(\to\) 82.07 \\
GLM & Claude identity & 67.41 \(\to\) 67.06 \(\to\) 68.17
  & 73.37 \(\to\) 73.69 \(\to\) 73.37 & 78.00 \(\to\) 76.89 \(\to\) 78.37 \\
GLM & knowledge cutoff & 67.41 \(\to\) 66.55 \(\to\) 66.98
  & 73.37 \(\to\) 72.52 \(\to\) 72.48 & 78.00 \(\to\) 74.68 \(\to\) 76.89 \\
GLM & context window & 67.41 \(\to\) 68.17 \(\to\) 67.75
  & 73.37 \(\to\) 74.23 \(\to\) 73.48 & 78.00 \(\to\) 78.37 \(\to\) 78.56 \\
\(\mathrm{GLM}_{\mathrm{think}}\) & Llama identity & 61.86 \(\to\) 59.90 \(\to\) 61.26
  & 83.26 \(\to\) 83.90 \(\to\) 82.83 & 86.51 \(\to\) 86.14 \(\to\) 86.69 \\
\(\mathrm{GLM}_{\mathrm{think}}\) & knowledge cutoff & 61.86 \(\to\) 58.79 \(\to\) 60.58
  & 83.26 \(\to\) 82.12 \(\to\) 79.91 & 86.51 \(\to\) 82.44 \(\to\) 82.81 \\
\(\mathrm{GLM}_{\mathrm{think}}\) & context window & 61.86 \(\to\) 57.94 \(\to\) 62.12
  & 83.26 \(\to\) 74.36 \(\to\) 80.41 & 86.51 \(\to\) 72.64 \(\to\) 79.30 \\
\hline
Qwen & adult-content refusal & 64.68 \(\to\) 55.29 \(\to\) 56.48
  & 73.63 \(\to\) 60.89 \(\to\) 61.12 & 82.07 \(\to\) 55.45 \(\to\) 58.78 \\
GLM & adult-content refusal & 67.41 \(\to\) 61.18 \(\to\) 63.14
  & 73.37 \(\to\) 58.07 \(\to\) 63.47 & 78.00 \(\to\) 55.08 \(\to\) 60.63 \\
\(\mathrm{GLM}_{\mathrm{think}}\) & adult-content refusal & 61.86 \(\to\) 40.61 \(\to\) 59.13
  & 83.26 \(\to\) 39.03 \(\to\) 76.78 & 86.51 \(\to\) 62.85 \(\to\) 81.70 \\
\hline
\end{tabular}
\end{table}

\section{Directional-Ablation Baseline Protocol}
\label{app:arditi-protocol}

The baseline follows \citet{arditi2024refusal} with the fitting data replaced by
L2W's probe set. The positive set is each model's Target-Semantic refusal
probes, six for Qwen and five for GLM in both configurations; the negative set
pairs each probe with a non-refusal writing request of the same form and
domain. Each candidate direction is the difference between the two sets' mean
residual activations at one layer and one assistant-template suffix position;
sweeping every layer outside the final 20\% and every suffix position (nine for
Qwen, two for GLM) gives the candidate pool. With thinking enabled, each prompt
generates its own trace (Appendix~\ref{app:glm-thinking}), so suffix positions
no longer align across prompts; the candidate position is fixed at the token
that closes the trace, and the pool sweeps layers only.

Selection keeps the official gates and thresholds. A candidate must pass two
gates on the negative prompts: ablating it must leave their behavior nearly
unchanged (next-token KL divergence at most 0.1), showing the direction
carries nothing else the model needs, and adding it must induce refusal on
them, showing the direction indeed drives refusal. Among eligible candidates,
the one whose ablation leaves the least refusal on the positive prompts is 
selected. Qwen has five eligible candidates
and GLM three in each configuration; the Qwen and thinking-disabled GLM
directions come from layer 19 and the final template position, and the
thinking-enabled GLM direction from layer 29 at the trace-closing token. The
gates score refusal from next-token probabilities only;
all reported outcomes use the same LLM judge as the other experiments. The 
selected direction is projected out of the residual stream at every layer and
token position. Table~\ref{tab:arditi-refusal} compares the resulting
directional-ablation baseline with L2W on the three refusal runs.

\begin{table}[h]
\centering
\caption{Directional ablation compared with L2W on the three adult-content
refusal runs.}
\label{tab:arditi-refusal}
\scriptsize
\setlength{\tabcolsep}{3pt}
\begin{tabular}{llrrrrrr}
\hline
Model & Method & \(\mathcal P_{\mathrm{sem}}\) & \(\mathcal P_{\mathrm{cap}}\)
  & \(\mathcal P_{\mathrm{held}}\) & ARC-Challenge & MMLU & IFEval \\
\hline
Qwen & Baseline & 6/6 & 10/10 & 10/10 \(\cdot\) 0/10 & 64.68 & 73.63 & 82.07 \\
Qwen & Directional ablation & 1/6 & 10/10 & 1/10 \(\cdot\) 10/10 & 64.76 & 73.19
  & 80.96 \\
Qwen & L2W post-removal & 0/6 & 9/10 & 0/10 \(\cdot\) 10/10 & 55.29 & 60.89
  & 55.45 \\
Qwen & L2W post-restoration & 0/6 & 10/10 & 0/10 \(\cdot\) 10/10 & 56.48 & 61.12
  & 58.78 \\
\hline
GLM & Baseline & 5/5 & 10/10 & 10/10 \(\cdot\) 0/10 & 67.41 & 73.37 & 78.00 \\
GLM & Directional ablation & 0/5 & 10/10 & 1/10 \(\cdot\) 10/10 & 68.26 & 72.94
  & 78.93 \\
GLM & L2W post-removal & 0/5 & 3/10 & 0/10 \(\cdot\) 7/10 & 61.18 & 58.07
  & 55.08 \\
GLM & L2W post-restoration & 0/5 & 10/10 & 0/10 \(\cdot\) 10/10 & 63.14 & 63.47
  & 60.63 \\
\hline
\(\mathrm{GLM}_{\mathrm{think}}\) & Baseline & 5/5 & 10/10 & 10/10 \(\cdot\) 0/10
  & 61.86 & 83.26 & 86.51 \\
\(\mathrm{GLM}_{\mathrm{think}}\) & Directional ablation & 0/5 & 10/10
  & 0/10 \(\cdot\) 10/10 & 62.80 & 83.90 & 84.66 \\
\(\mathrm{GLM}_{\mathrm{think}}\) & L2W post-removal & 0/5 & 9/10
  & 0/10 \(\cdot\) 6/10 & 40.61 & 39.03 & 62.85 \\
\(\mathrm{GLM}_{\mathrm{think}}\) & L2W post-restoration & 0/5 & 10/10
  & 0/10 \(\cdot\) 10/10 & 59.13 & 76.78 & 81.70 \\
\hline
\end{tabular}
\end{table}

\section{Text-to-Image Post-Editing Protocol}
\label{app:flux-protocol}

The composition experiment of Section~\ref{sec:sensitive-content-excision} fixes
the prompt, seed, initial latent, resolution, four-step sampler, and guidance
scale. We fit J-lenses targeting outputs of the text encoder's 9th, 18th, and
27th layers using 1,000 WikiText-103 prompts. L2W intervenes only on physical
attention heads and MLP channels in the text encoder; the image transformer and
decoder remain unchanged. The CLIP contrast registers two labels, clothed and
nude: the similarity margin between the rendered image and the two labels
certifies whether the target is present, and the margin's gradient into the text
encoder gives J-lens attribution its scoring direction.

\end{document}